\documentclass[runningheads]{llncs}
 
\usepackage{eccv}

\usepackage{eccvabbrv}

\usepackage{graphicx}
\usepackage{booktabs}
\usepackage{algorithm}
\usepackage{algorithmic}
\usepackage{multirow}
\usepackage{threeparttable}

\usepackage{amssymb}
\usepackage{xcolor}
\usepackage{caption}
\usepackage{amsmath}
\usepackage{adjustbox}

\usepackage{cancel}
\usepackage{pifont}
\newcommand{\xmark}{\ding{55}}%

\usepackage{epigraph}
\definecolor{citecolor}{HTML}{0071bc}
\definecolor{ourscolor}{HTML}{c2d1e5}
\usepackage{colortbl}  
\newcolumntype{a}{>{\columncolor{ourscolor}}c}

\definecolor{darkcandyapplered}{rgb}{0.64, 0.0, 0.0}
\definecolor{burgundy}{rgb}{0.5, 0.0, 0.13}
\definecolor{carnelian}{rgb}{0.7, 0.11, 0.11}

\definecolor{tabhighlight}{HTML}{ffffff} 
\definecolor{tabhighlightric}{HTML}{ffffff} 

\usepackage[misc]{ifsym}

\makeatletter
\def\blfootnote{\gdef\@thefnmark{}\@footnotetext}
\makeatother
\usepackage[accsupp]{axessibility}  

\usepackage{hyperref}

\usepackage{orcidlink}

\begin{document}

\title{UniFS: Universal Few-shot Instance Perception with Point Representations} 


\author{
    Sheng Jin\inst{1,4}\thanks{Equal contribution. \quad \Letter~Corresponding authors.}\orcidlink{0000-0001-5736-7434} \and
    Ruijie Yao\inst{2,4 \star} \orcidlink{0009-0007-4736-9453} \and
    Lumin Xu\inst{3}\orcidlink{0000-0003-2125-2760} \and
    Wentao Liu\inst{4}\orcidlink{0000-0001-6587-9878} \textsuperscript{\Letter} \and
    Chen Qian\inst{4}\orcidlink{0000-0002-8761-5563} \\
    Ji Wu\inst{2} \orcidlink{0000-0001-6170-726X}  \and
    Ping Luo\inst{1,5}\orcidlink{0000-0002-6685-7950} \textsuperscript{\Letter}
}

\authorrunning{S. Jin et al.}

\institute{$^{1}$ The University of Hong Kong \quad
$^{2}$ Tsinghua University \quad
$^{3}$ The Chinese University of Hong Kong \quad
$^{4}$ SenseTime Research and Tetras.AI \quad
$^{5}$ Shanghai AI Laboratory \\
\email{js20@connect.hku.hk, yrj21@mails.tsinghua.edu.cn}}

\maketitle
\begin{abstract}
Instance perception tasks (object detection, instance segmentation, pose estimation, counting) play a key role in industrial applications of visual models. 
As supervised learning methods suffer from high labeling cost, few-shot learning methods which effectively learn from a limited number of labeled examples are desired. 
Existing few-shot learning methods primarily focus on a restricted set of tasks, presumably due to the challenges involved in designing a generic model capable of representing diverse tasks in a unified manner.
In this paper, we propose UniFS, a universal few-shot instance perception model that unifies a wide range of instance perception tasks by reformulating them into a dynamic point representation learning framework. 
Additionally, we propose Structure-Aware Point Learning (SAPL) to exploit the higher-order structural relationship among points to further enhance representation learning. 
Our approach makes minimal assumptions about the tasks, yet it achieves competitive results compared to highly specialized and well optimized specialist models. Codes and data are available at \url{https://github.com/jin-s13/UniFS}.

\keywords{Few-Shot Learning \and Unified Vision Model}
\end{abstract}

\section{Introduction}

The field of instance perception, also known as object-centric learning, has made significant progress in recent years, with notable advancements. This paper primarily concentrates on four fundamental sub-tasks: object detection, instance segmentation, pose estimation\footnote[1]{In this paper, we use ``keypoint localization'' and ``pose estimation'' interchangeably.}, and object counting.
Traditional approaches in this domain typically employ supervised learning methods, which necessitate enormous and costly annotated data, resulting in high costs. In contrast, there is a growing interest in few-shot learning methods that can leverage prior knowledge to handle novel concepts using only a small number of labeled samples.

Existing few-shot learning methods for instance perception are typically developed for a single sub-task or a subset of sub-tasks. Unifying different instance perception tasks into a general model meets significant challenges due to variations in data sources, feature granularity, and output structures.
(a) From a data perspective, images in different instance perception tasks and datasets exhibit diverse characteristics and inherent scale variances. For example, object detection datasets~\cite{lin2014microsoft} often consist of scene images with multiple interacting objects, while pose estimation datasets~\cite{xu2022pose} typically contain center-cropped images featuring a single object.
(b) Regarding feature learning, different tasks require the extraction of relevant features at various levels of granularity. For instance, object detection prioritizes global semantic features, instance segmentation relies on fine-grained semantic features, and pose estimation necessitates both fine-grained semantic and localization features.
(c) From the perspective of output, the outputs of different instance perception tasks have distinct structures. For example, the object detection task uses the coordinates, width and height of bounding box, while the pose estimation task employs Gaussian heatmaps for keypoint localization.
These challenges hamper the development of multi-task few-shot learning methods that can comprehensively address a wide range of instance perception tasks in a unified manner.

\begin{figure}[t]
\centering
    \includegraphics[width=0.8\textwidth]{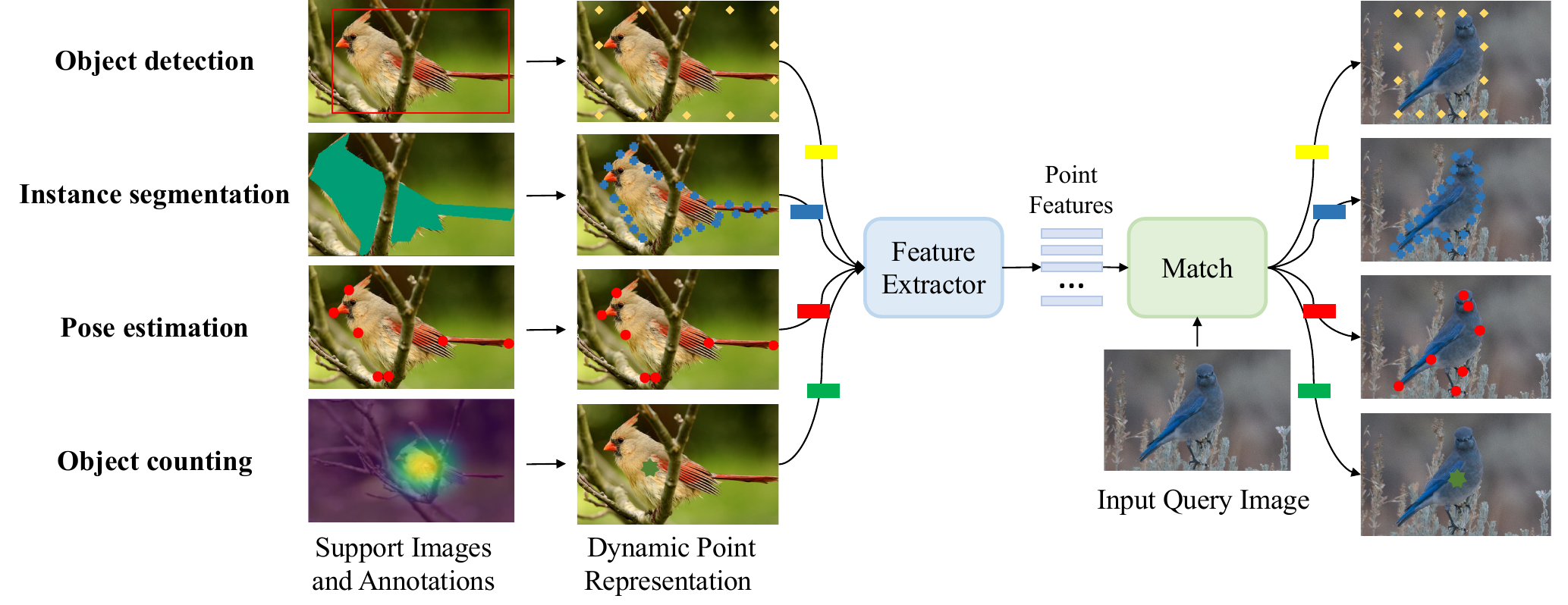}
\caption{UniFS utilizes a dynamic point representation learning paradigm to merge various task output spaces into a set of multiple points. With the provision of few-shot support images and annotated points from different tasks, UniFS can seamlessly produce corresponding point outputs on the query set.
}
\label{fig:instance_perception}
\vspace{-5mm}
\end{figure}

Nevertheless, building a powerful generalist model that learns to solve multiple tasks~\cite{lake2013one} from only a few scraps of information is an ultimate goal in the AI community. 
Current approaches often adopt task-specific formulations, which involve separate learnable components and training objectives for each task. Such designs trend to be complicated and less scalable as the number of tasks increases. Although there have been recent works on generalist few-shot learners~\cite{kim2023universal,sun2023uniap,bohdal2023meta}, they primarily focus on dense map prediction tasks, thus,  cannot distinguish individual object instances. 
The problem of universal few-shot learning for instance perception tasks remains largely unexplored.

In this paper, we propose UniFS, that is a universal model with a task-agnostic structure designed to unify a wide range of few-shot instance perception tasks. As shown in Fig.~\ref{fig:instance_perception}, the key idea is to reformulate diverse instance perception tasks into a generalized point representation learning paradigm. By providing user-defined point annotations in the support set, the model is prompted to learn task-specific points for each query image. For example, in object detection, points are placed along the bounding box edges, while for instance segmentation, points are distributed along the instance mask contour. This unified formulation of point learning brings several benefits:
(1) \textbf{Task-agnostic architecture:} The model uses a unified representation that supports diverse task types and breaks the scaffolds of complex task-specific customization. 
(2) \textbf{Parameter efficiency:} Parameters are fully shared among tasks, reducing redundant computation when handling multiple tasks simultaneously.
(3) \textbf{Task generalization:} The model can generalize to new tasks with minimal knowledge of the underlying mechanism.
(4) \textbf{Knowledge sharing:} Joint training on diverse tasks enables knowledge sharing among tasks, enhancing overall performance.

Furthermore, we propose a novel structure-aware point learning objective to address the limitations of traditional L1/L2 losses in point learning. By considering the relationship between the center point and its neighboring points, our objective leverages contextual information and spatial dependencies within the local region. This leads to more accurate predictions by better capturing spatial relationships and structural patterns among points. Additionally, the inclusion of structural relationships provides regularization, enabling the model to handle noisy inputs and outliers more effectively, resulting in robust and stable training.

We address the lack of a comprehensive benchmark for evaluating multiple instance perception tasks by introducing the COCO-UniFS dataset. Derived and re-organized from existing datasets, COCO-UniFS encompasses tasks including object detection, instance segmentation, pose estimation, and object counting. This unified dataset serves as a platform for developing and assessing multi-task few-shot learning methods, facilitating model evaluation and generalization across various instance perception tasks. Our approach achieves competitive results on COCO-UniFS benchmark, with minimal task assumptions, compared to highly specialized task-specific models.

Our work makes the following key contributions:

\begin{itemize}
    \item \textbf{Pioneering Universal Few-Shot Instance Perception:} To the best of our knowledge, we are the first to tackle the problem of universal few-shot learning for instance perception tasks. By introducing the general and unified vision problem and benchmark, we hope to inspire the research community to explore this promising direction.

    \item \textbf{A Simple and Effective Baseline:} We propose UniFS, a simple yet effective baseline approach that unifies multiple instance perception tasks under a few-shot point representation learning paradigm. It employs a unified architecture for all tasks and makes minimal assumptions about the specific characteristics of each task.

    \item \textbf{Structure-Aware Point Learning:} We introduce a novel structure-aware point learning objective that captures the higher-order structural relationships among points, enhancing the model's ability to understand and exploit the spatial dependencies among points.

\end{itemize}

\section{Related Works}

\subsection{Few-Shot Learning} 

\textbf{Few-Shot Learning Methods.} 
Few-shot learning aims to imitate humans’ generalization ability that can efficiently learn novel concepts from a few examples. It has gained attention in recent years, with approaches falling into meta-learning, metric-learning, and transfer learning (or finetuning) paradigms. 
Meta-learning based methods~\cite{wang2019meta,kang2019few,yan2019meta,xiao2020few,fan2020few,li2021transformation,li2021beyond,zhu2021semantic,hu2021dense,han2022few} intent to learn efficient parameter updating rules or better parameter initialization strategies.
Metric-learning based methods~\cite{koch2015siamese,snell2017prototypical,sung2018learning,vinyals2016matching} focus on obtaining a generalizable embedding metric space for pairwise similarity comparisons. 
Transfer learning based methods~\cite{sun2021fsce,cao2021few,qiao2021defrcn,gao2022decoupling,wang2020frustratingly,wu2020multi} leverage pre-training on base classes and fine-tuning on novel classes. Our UniFS belongs to the transfer learning category.

\textbf{Few-Shot Learning Tasks.} 
Few-shot learning has extended beyond classification to applications like object detection \cite{sun2021fsce,cao2021few,qiao2021defrcn,wang2019meta,kang2019few,xiao2020few,fan2020few,li2021transformation,li2021beyond,zhu2021semantic}, instance segmentation \cite{ganea2021incremental,gao2022decoupling,fan2020fgn,michaelis2018one,yan2019meta}, pose estimation \cite{xu2022pose,hirschorn2023pose}, and object counting \cite{ranjan2021learning,you2023few}. However, existing approaches tend to be task-specific, resulting in diverse models and datasets. These specialized methods hinder integration into a unified framework. For example, DeFRCN \cite{qiao2021defrcn} uses a two-stage detector for few-shot object detection, while SAFECount \cite{you2023few} employs a single-stage approach for few-shot counting. Although some methods handle pairs of tasks like object detection and instance segmentation \cite{ganea2021incremental,gao2022decoupling,yan2019meta}, they rely on task-specific components. In contrast, our UniFS model minimizes task-specific designs through a unified architecture. While recent efforts focus on generalist few-shot learners for tasks like semantic segmentation \cite{kim2023universal,sun2023uniap,bohdal2023meta}, they overlook instance perception tasks. To the best of our knowledge, universal few-shot learning for instance perception tasks remains unexplored, which is the main focus of this paper.

\textbf{Few-Shot Learning Benchmarks.}
Most existing few-shot learning benchmarks are designed for specific computer vision tasks, such as image classification~\cite{lake2015human,vinyals2016matching}, semantic segmentation~\cite{li2020fss}, object detection~\cite{lin2014microsoft}, pose estimation~\cite{xu2022pose}, and object counting~\cite{ranjan2021learning}. Due to the lack of a unified benchmark dataset, it is hard to evaluate and develop generalist few-shot learners capable of solving diverse vision tasks. 
In this paper, we introduce the COCO-UniFS benchmark dataset, which extends existing datasets and covers multi-task few-shot learning of instance perception tasks. Existing datasets for multi-task few-shot learning, such as Taskonomy~\cite{zamir2018taskonomy,kim2023universal} and Meta Omnium~\cite{bohdal2023meta}, do not target at instance perception tasks like object detection and instance segmentation. Moreover, these datasets are often combinations of several sub-datasets, making it challenging to analyze the impact factors involved.

\subsection{Unified Vision Models} 

\textbf{Unified Learning Paradigms.}
Several works have focused on general representation learning for foundation vision models~\cite{ghiasi2021multi,shao2021intern,wang2022omnivl,zhu2022uni,ci2023unihcp,tang2023humanbench,zeng2022not,chen2023beyond,hong2022versatile,zhang2024when}.
However, these models typically require costly task-specific fine-tuning to achieve optimal performance on specific downstream tasks. In contrast, UniFS differs from them, as it aims to address multi-task few-shot learning.

\textbf{Unified Task Representations.} Efforts have been made to unify task output representations for multiple vision tasks.
\textbf{Dense representations.} Some studies~\cite{wang2023images,kim2023universal} employ dense image-like representations, as seen in Painter~\cite{wang2023images}, facilitating tasks like panoptic segmentation and depth estimation. However, they lack versatility in addressing core vision tasks such as object detection and instance segmentation.
\textbf{Text representations.} Other approaches~\cite{chen2021pix2seq,chen2022unified,yang2022unitab,wang2022ofa,lu2022unified} leverage text representations inspired by NLP's sequence-to-sequence modeling. Despite their potential, they suffer from slow inference speeds and overlook category-agnostic multi-instance pose estimation.
\textbf{Point representations} are prevalent in various vision tasks, including object detection~\cite{law2018cornernet,duan2019centernet}, segmentation~\cite{xie2020polarmask}, and pose estimation~\cite{tian2019directpose,nie2019single,wei2020point}. While some works attempt to unify tasks using point representations, they often train and evaluate models in a single-task manner. In this paper, we propose a universal generalist model unifying few-shot instance perception tasks through point representations.

\textbf{Unified Model Architectures.}
Designing a unified model capable of handling different tasks in a unified manner is considered a step toward general intelligence. 
Some attempts~\cite{he2017mask,jin2023you,yan2022towards,jin2020whole,xu2022zoomnas} have been made to design unified model architectures for visual tasks, but they often rely on task-specific heads to perform different tasks, and the interactions among different tasks are less studied. 
UniHCP~\cite{ci2023unihcp} minimizes task-specific parameters for knowledge sharing but still requires task-specific queries and a task-guided interpreter. Additionally, it is designed specifically for human-centric tasks. In our work, we take a further step by unifying few-shot instance perception tasks with a fully shared model architecture, without any task-specific designs.

\section{COCO-UniFS Benchmark}
This work targets at universal few-shot instance perception that uniformly handles multiple tasks for arbitrary categories by providing a few visual examples and the corresponding annotations from that category. 
In order to address the task of universal few-shot instance perception, we introduce the COCO-UniFS benchmark. This benchmark is built upon several existing datasets, including MSCOCO~\cite{lin2014microsoft} and MISC~\cite{sun2023misc210k}, aiming to provide a high-quality and high-diversity dataset for model training and evaluating this challenging task. 
We have made efforts to ensure unification for dataset annotation, dataset split, and evaluation protocols, enabling fair comparisons between models and methods.

\subsection{Task Definition} 
We define the task of universal few-shot instance perception as follows: 
given a few reference images of a novel object category and the corresponding annotations (\eg box, mask, keypoint, or center point), this task requires the model to find corresponding labels of all the instances belonging to this category in the query image. 
The reference image shows a cropped single object instance of the category that is to be analyzed, while the query image shows an entire visual scene containing many objects. 
Note that we assume that the task information (\eg detection or segmentation) is implicitly embedded in the given annotations, and is not explicitly provided. This makes it possible to define your own ``novel'' task by offering varying annotations in the support images.

\subsection{Data Annotation} 
The COCO-UniFS benchmark provides dense annotations for four fundamental few-shot computer vision tasks: object detection, instance segmentation, pose estimation, and object counting.
The annotations for object detection and instance segmentation are directly taken from the MSCOCO dataset~\cite{lin2014microsoft}, which provides bounding box and per-instance segmentation mask annotations for 80 object categories. 
For pose estimation, we extend the MSCOCO dataset by adding instance-level keypoint annotations for 34 object categories from the MISC dataset~\cite{sun2023misc210k}. The MISC dataset was originally designed for multi-instance semantic correspondence, and we adapted it to fit the few-shot pose estimation task. 
The keypoints are defined on the 3D prototype model, encompassing contour points (\eg head top), skeletal joints (\eg wrists), and other distinctive feature points (\eg eyes). 
We manually add keypoint semantic descriptions with their symmetry relationship for each object category. 
Objects with ambiguous or inconsistent keypoint definitions, insufficient keypoint numbers, or large morphology variations are filtered out to ensure annotation accuracy.
Note that the keypoint definition (even the number of keypoints) varies from category to category. 
For object counting, each object is annotated with a point at the center of its bounding box, following the method in prior datasets~\cite{ranjan2021learning}.

\subsection{Dataset Split}
To evaluate few-shot generalization to unseen categories, the available object categories should be split into a training set and a non-overlapping test set.
Previously, due to the independent development of various few-shot learning tasks, different conventions exist for dataset splits across research areas.
To facilitate multi-task few-shot learning and ensure consistency, we unify the dataset split for all tasks in the COCO-UniFS benchmark.
The dataset split follows the convention used in COCO-inst~\cite{lin2014microsoft} for few-shot object detection and instance segmentation. The 20 overlapping categories with PASCAL VOC 2012 are novel classes, while the remaining 60 categories are base classes.

\subsection{Evaluation Scenarios and Metrics} 

\textbf{Evaluation Scenarios.}
We establish two evaluation scenarios for COCO-UniFS:
(1) \textbf{Seen-task evaluation}:  In the seen-task few-shot evaluation, the evaluation tasks are seen and trained during the base training phase.
This evaluation scenario focuses on the tasks of object detection, instance segmentation, and pose estimation. 
(2) \textbf{Unseen-task evaluation}: This evaluation scenario is designed to evaluate novel task generalization, specifically for the task of object counting. {Note that the evaluation categories are unseen in both evaluation scenarios.}

\textbf{Evaluation Metric.}
For object detection, instance segmentation, and pose estimation, we use the COCO official metric and report the Average Precision (AP) with IoU (Intersection over Union) threshold ranging from 0.5 to 0.95 on the novel classes, where box IoU, mask IoU, and Object Keypoint Similarity (OKS) based IoU are applied respectively. 
Note that for pose estimation, we only consider the keypoints that exist both in at least one support image and in the query image during metric calculation.
For object counting, we use Mean Squared Error (MSE) as the main evaluation metric. Previous works~\cite{ranjan2021learning,yang2021class} calculate the average over the images, instead of categories. To make the evaluation of object counting consistent with other tasks, we opt to calculate the mean MSE by averaging the MSE values over all the evaluation classes.

\subsection{Uniqueness} 
The COCO-UniFS dataset stands out due to several unique properties:
(1) \textbf{Comprehensiveness:} The COCO-UniFS dataset covers four fundamental instance perception tasks: object detection, instance segmentation, pose estimation, and object counting.  
With 100,000+ valid images, it features diverse lighting, resolutions, and object categories.
(2) \textbf{Annotation Completeness:}  
The COCO-UniFS dataset differs from previous benchmarks by offering complete annotations for all tasks, eliminating the need for multiple sub-datasets. This comprehensive annotation simplifies algorithm development for handling multiple tasks concurrently.
(3) \textbf{Domain Uniformity:} 
Existing multi-task benchmark datasets such as Meta Omnium~\cite{bohdal2023meta} and Taskonomy~\cite{zamir2018taskonomy,kim2023universal} are composed of multiple sub-datasets from different visual domains, ranging from natural images to medical and industrial images. This introduces additional domain factors that affect model learning and evaluation. 
In contrast, the COCO-UniFS dataset maintains domain uniformity,  where all images and tasks are within the same data domain. This simplifies the development and analysis of multi-task few-shot learners, allowing researchers to concentrate on instance perception challenges without the interference of domain shifts.
(4) \textbf{Unseen Task Generalization:} 
The COCO-UniFS benchmark introduces an unseen-task evaluation scenario for object counting, enabling researchers to assess model generalization to new tasks unseen during training. These features make the COCO-UniFS dataset a valuable asset for advancing multi-task few-shot instance perception research.

\section{Methodology}

\begin{figure}[t]
\centering
    \includegraphics[width=0.98\textwidth]{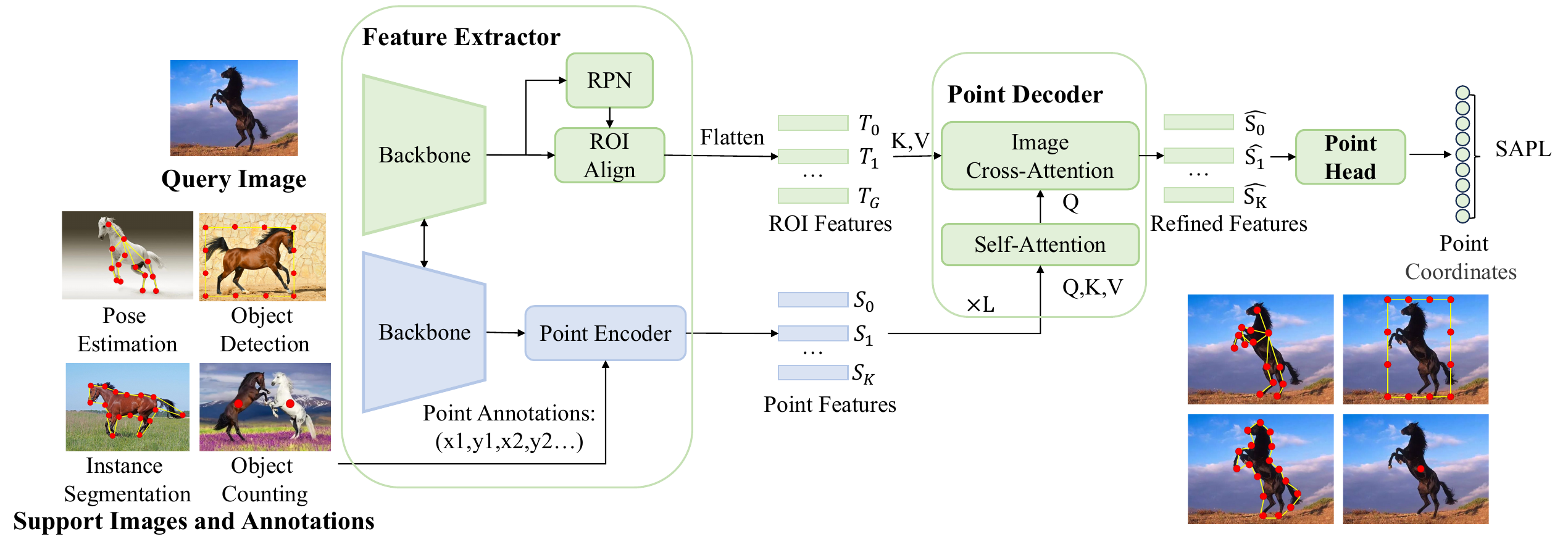}
\caption{Overview of UniFS. UniFS adopts a dynamic point representation learning paradigm to unify different task output spaces into a set of multiple points.}
\label{fig:overview}
\end{figure}

\subsection{Overview}

Our goal is to develop a single unified model capable of addressing a diverse set of few-shot learning tasks without the need for task-specific algorithm customizations (\eg output representations and loss functions) or model parameters (\eg separate architectures and task heads). 
The overall pipeline of our proposed UniFS is presented in Fig.~\ref{fig:overview}.
Our focus lies in unifying task representation, model architecture, and learning paradigms.

\subsection{Unified Task Representation}
Traditionally, instance perception tasks, such as image classification~\cite{lake2015human,vinyals2016matching,yao2024gkgnet}, object detection~\cite{zhang2022dino,zhang2024when}, instance segmentation~\cite{he2017mask,xie2020polarmask,li2023mask}, and pose estimation~\cite{xiao2018simple,jin2019multi,jin2020differentiable,xu2021vipnas,xu2022zoomnas,jiang2022posetrans}, each have their distinct representations,  posing challenges for simultaneous handling.
To achieve unified instance perception, we propose a point-based formulation for various tasks, offering a unified representation as follows:
\textbf{Object Detection.}
Object detection task is a fundamental computer vision task in which the goal is to obtain the bounding box level location of objects. 
we utilize 16 points to depict a bounding box, which are uniformly sampled across the box. 
\textbf{Instance Segmentation.}
Instance segmentation requires pixel-wise masks, so we employ 32 points to outline the boundary of each target instance. In the COCO dataset, segmentation masks have been annotated using either clockwise or counterclockwise polygons.
Following the methodology outlined in Deep Snake~\cite{peng2020deep}, we uniformly add or remove points from the initial contour until the desired number of points is reached. We ensure a consistent clockwise orientation and designate the point situated at the leftmost edge of the contour as the starting point.
\textbf{Pose Estimation.}
Pose estimation detects the coordinates of semantic keypoints. Our point-based representation naturally aligns with this task, with each point corresponding to a target keypoint. This facilitates estimating keypoints based on definitions in the support image, accommodating various object categories.
\textbf{Object Counting.}
UniFS adopts a center point-based approach by predicting the center point of each target object. The center point is defined as the midpoint of the object's bounding box. By counting the center points for a specific category, we can determine the number of objects in that category.
A detailed ablation of the number of points is provided in supplementary Table S1.

\subsection{Unified Model Architecture}

The UniFS model consists of three main components: the feature extractor, the point decoder, and the point head. These components work together to handle different instance perception tasks within a unified framework.

\textbf{Feature Extractor.} 
The feature extractor is proposed to extract features from the support and query images. We utilize an ImageNet~\cite{krizhevsky2017imagenet} pre-trained ResNet-101~\cite{he2016deep} backbone as the feature extractor, which is shared between both the support image and the query image. Given the support image $I_S$, the feature extractor $\mathcal{F}$ generates the support image features $F_S = \mathcal{F}(I_S)$. 
Sinusoidal Position embeddings~\cite{vaswani2017attention} are then added to the support image features.
Given ground-truth points $[(x_1, y_1), (x_2, y_2), ..., (x_K, y_K)]$ for the support sample, we obtain the point features $\{S_i\}_{i\in[1, K]}$ by extracting the features via bi-linear interpolation from the corresponding location of the support image features $F_S$.
For the query image $I_Q$, the shared feature extractor is applied to extract the query image features $F_Q = \mathcal{F}(I_Q)$. Then, a Region Proposal Network (RPN) generates candidate regions, and RoI Align is applied to pool each region proposal feature into a sequence of flattened query object features $\{T_i\}_{i\in[1,G]}$, where $G$ is the number of flattened RoI features for each given region proposal.
Following Faster R-CNN~\cite{ren2015faster}, the MLP-based object classification head is applied to {the proposal features that are obtained by global pooling of region proposal features} to predict categories of the objects.

\textbf{Point Decoder.} 
To capture the correlative dependence between the support point set and the query point set, we introduce transformer-based attention mechanisms to enhance the representations of point features $\{S_i\}_{i\in[1, K]}$. The point decoder module consists of $L$ transformer decoder layers. Each layer contains two main components: Self-Attention and Cross-Attention. 
\emph{Self-Attention.} The self-attention layer~\cite{vaswani2017attention} enables the exchange of information among the query points, and obtains the refined point features $\{S_i^{'}\}_{i\in[1, K]}$. By allowing the query point features to interact with each other and aggregating these interactions using attention weights, the contextual meaning of the current point sequence, particularly the information relevant to the task, is better modeled.
\emph{Cross-Attention.} Then point features interact with the query image features to align the feature representations and mitigate the representation gap. In this case, a cross-attention layer~\cite{carion2020end} is used to aggregate useful information between the support point and query image. The support point features $\{S_i^{'}\}_{i\in[1, K]}$ are used as queries, while the flattened query RoI features $\{T_i\}_{i\in[1, G]}$ of each region proposal serve as the keys and values. 
The channel dimension of the query RoI features is reduced to match the channel dimension of the support point features, and sinusoidal position embeddings~\cite{vaswani2017attention} are added to the query RoI features. A feed-forward network (FFN) and layer normalization (LN) are also included following common practice.
Finally, this process yields an enhanced group of point features $\{\widehat{S_i}\}_{i\in[1, K]}$ for each set of RoI features. The computation in point decoder can be formulated as follows:
\begin{equation}
S_i^{'}=LN(SelfAttn(S_i)), \quad
\widehat{S_i}=LN(FFN(LN(CrossAtten(S_i^{'},T_i))))
\end{equation}

\textbf{Point Head.} 
Given the refined point embeddings $\{\widehat{S_i}\}$, we utilize a multi-layer perceptron (MLP) to transform them into the point coordinates. Rather than directly predicting absolute coordinates, we infer the offset $(\Delta x_i\, \Delta y_i)$ from the center of RPN anchors. Finally, the scattered point coordinates $(P_{xi}, P_{yi})$ are generated by adding the offset prediction to the proposal anchor bounding box $A: (A_{cx}, A_{cy}, A_w, A_h)$: 
\begin{equation}
P_{xi} = A_{cx} + \Delta x_i \times A_w, \quad
P_{yi} = A_{cy} + \Delta y_i \times A_h.
\end{equation}

\subsection{Unified Learning Paradigm}
In contrast to previous approaches that use different training objectives and loss functions for different tasks, we propose a unified learning paradigm for all available tasks. This approach enables us to train the model jointly on multiple tasks, leading to improved efficiency and performance.

\begin{figure}[t]
\centering
    \includegraphics[width=0.9\textwidth]{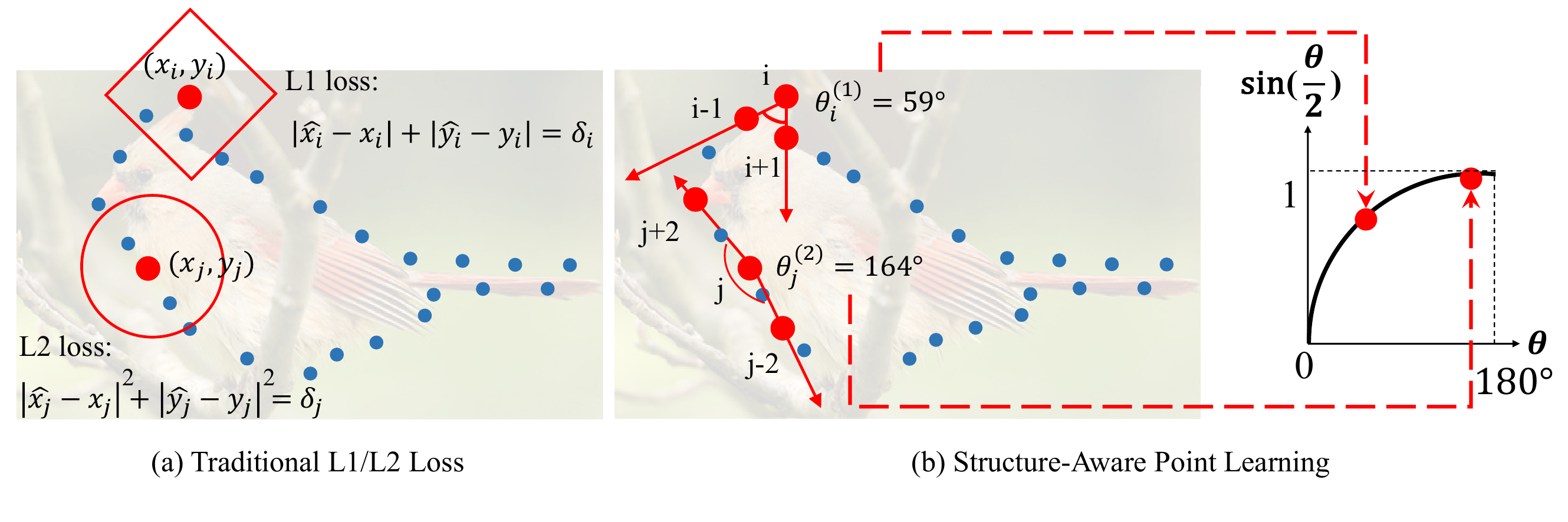}
\caption{\textbf{Structure-Aware Point Learning (SAPL)}: (a) Traditional L1/L2 loss focuses on individual point error. (b) SAPL integrates structural relationships among points by supervising the angle between each point and its neighboring points.
}
\label{fig:sapl}
\end{figure}

\textbf{Structure-Aware Point Learning (SAPL).}
Fig.~\ref{fig:sapl} illustrates the differentiation between conventional point-level loss and our structure-aware point learning loss. While L1/L2 loss functions focus solely on individual point-level error, our proposed Structure-Aware Point Learning (SAPL) incorporates the inter-dependency between a specific point and its neighboring points, thus capturing the intrinsic structural information within the point sequence. As depicted in Fig.~\ref{fig:sapl}(a), while predicted points on the diamond shape result in identical L1 losses, those on the circle yield the same L2 losses. 
This introduces ambiguity in the model's learning process. However, integrating SAPL helps mitigate such ambiguity, thus enhancing the representation learning.

Specifically, we additionally impose constraints on the structural information by supervising the angles \(\theta_i^{(n)}\) between each point and its adjacent n-hop points. Here, \(\theta_i^{(n)}\) represents the angle formed by points \(i-n\), \(i\), and \(i+n\). As illustrated in Fig.~\ref{fig:sapl}(b), the angle \(\theta_i^{(1)}\) formed by points \(i-1\), \(i\), and \(i+1\) for each point \(i\), while \(\theta_j^{(2)}\) demonstrates the 2-hop formulation. Incorporating higher-order information supervision helps reduce errors from point annotation noise, thereby enhancing robustness.
Additionally, we integrate a sine function \( \sin(\frac{\theta}{2}) \) into the loss function to amplify gradients at sharp peaks and attenuate gradients in flat regions. This enables the effective capture of detailed shape information.
Formally, the SAPL loss, integrating angles from 1-hop to N-hop, is defined as:
\begin{equation}
    L_{SAPL}(\widehat{P_i},P_i) = \frac{1}{N}\sum_{n=1}^{N}L_1(\sin(\frac{\widehat{\theta_i}^{(n)}}{2}), \sin(\frac{{\theta_i^{(n)}}}{2}))
\end{equation}
where \(\widehat{P_i}\) and \(P_i\) represent the predicted and ground-truth points, respectively. \(\widehat{\theta_i}^{(n)}\) and \(\theta_i^{(n)}\) denote their respective n-hop angles.
The final loss $L_{point}$ is computed as the sum of the L1 loss and SAPL loss for each point, given by:
\begin{equation}
    L_{point}(\widehat{P_i},P_i)=|\widehat{P_i}-P_i|+L_{SAPL}(\widehat{P_i},P_i)
\end{equation}

\section{Experiments}

\subsection{Implementation Details}

Experiments are conducted using Detectron2 (Wu et al., 2019). We employ the standard Faster R-CNN (Ren et al., 2015) with ResNet-101 (He et al., 2016) backbone. We set the number of point decoders \(L\) to 2 and configure the SAPL with an N-hop value of \(N = 2\).
For model training, we employed a two-stage transfer-learning approach: base class training and novel class fine-tuning. During base class training, models were jointly trained on object detection, instance segmentation, and pose estimation tasks simultaneously. Notably, the task of object counting was held out and not trained during this stage. Each batch randomly sampled different tasks. Due to the stochastic nature of task selection within each batch, employing a larger batch size was crucial for ensuring stable training. Thus, we set the batch size to 32 (4 per GPU * 8 GPUs), with a learning rate of 0.028, and utilized Stochastic Gradient Descent (SGD) to optimize our network, with a maximum of 55,000 iterations. 
For novel class fine-tuning, our model was fine-tuned with K-shots for every class, and the learning rate was reduced to 0.01 during few-shot fine-tuning. Following the previous work~\cite{gao2022decoupling}, all experimental results are averaged over 10 seeds.

\subsection{Baseline Methods.}
We compare our method to task-specific and multi-task few-shot baseline models. For few-shot object detection, we compare with several representative methods~\cite{yan2019meta,wang2020frustratingly,cao2021few,han2022few,zhang2022meta} including meta-learning based and metric-learning based methods.
For few-shot instance segmentation, we compare with a number of state-of-the-art models~\cite{yan2019meta,ganea2021incremental,ganea2021incremental,qiao2021defrcn,gao2022decoupling}. Especially, we also compare with the DETR-based models (Meta-DETR~\cite{zhang2022meta}) beyond the Faster R-CNN framework.
For few-shot object counting, we compare with both heatmap-based methods (CFOCNet~\cite{yang2021class}) and detection-based methods (MPSR~\cite{wu2020multi} and FSDetView~\cite{xiao2020few}). 
Please note that these models are trained with object counting tasks in the base-training stage, while our model performs unseen-task generalization. 
As no prior methods are developed for universal few-shot instance-perception tasks, we also adapt Mask-RCNN~\cite{he2017mask}, a well-known multi-task model, to our setup. Specifically, we add a heatmap based pose estimation head at the ROI of the Mask-RCNN. 
The baseline model initially trains on a mix of all base categories from the training set. During testing, it fine-tunes on support images of the new category. To address varying numbers of keypoints, the model outputs the maximum number across all categories, such as 52 on the COCO-UniFS dataset. Only valid keypoints for each category are supervised during training.

\subsection{Experiments on COCO-UniFS benchmark}
\label{sec:coco_results}

\subsubsection{Results on COCO-UniFS.}
We compare our method to task-specific and multi-task few-shot models on the COCO-UniFS dataset in Table~\ref{tab:coco_result}. All compared models adopt ResNet-101~\cite{he2016deep} as the backbone. 
We find that Our model significantly outperforms the multi-task generalist baseline, \ie Mask-RCNN-ft. Our approach achieves competitive results against state-of-the-art task-specific models, with only a slight disadvantage in 5-shot segmentation.
It is worth noting that our method has a better advantage, especially in extremely low-shot (K=1 shot), and thus it is very suitable for a few-shot scenario. 
\textbf{Unseen Task Generalization.} Our model also shows strong generalization capability when evaluated on the task of few-shot object counting (or center point localization). 
Please note that these baseline models are trained with object counting tasks in the base-training stage, while our model performs unseen-task generalization. Nevertheless, our UniFS outperforms them.

\begin{table*}[tb]
\caption{\textbf{Comparisons with task-specific and multi-task few-shot learners on the COCO-UniFS \texttt{val} set.} We report AP for object detection (Det.), instance segmentation (Seg.), and pose estimation (Kpt.), and report MSE for object counting (Cnt.). The results are averaged over all 10 seeds for fair comparisons. $\uparrow$ means the higher the better, while $\downarrow$ means the lower the better. $\dagger$ means seen-task evaluation, while $\ddagger$ means unseen-task generalization.
}
\centering
\small
\scalebox{0.75}{
\setlength{\tabcolsep}{10pt}
\begin{tabular}{l c c c c c c | c c}
\toprule
\multirow{2}{*}{Model} & \multicolumn{2}{c}{Det. AP $\uparrow$} &  \multicolumn{2}{c}{Seg. AP $\uparrow$} & \multicolumn{2}{c}{Kpt. AP $\uparrow$}  & \multicolumn{2}{|c}{Cnt. MSE $\downarrow$}\\
\cmidrule(r){2-3}
\cmidrule(r){4-5}
\cmidrule(r){6-7}
\cmidrule(r){8-9}
& K=1 & K=5  & K=1 & K=5  & K=1 & K=5  & K=1 & K=5   \\
\toprule
FRCN-ft~\cite{yan2019meta}  & 1.0 & 4.0      & \xmark & \xmark & \xmark & \xmark & \xmark & \xmark \\  
TFA~\cite{wang2020frustratingly}  & 4.4 & 7.7     & \xmark & \xmark & \xmark & \xmark & \xmark & \xmark \\  
FADI~\cite{cao2021few} &  5.7 & 10.1     & \xmark & \xmark & \xmark & \xmark & \xmark & \xmark \\ 
FCT~\cite{han2022few}  &  5.1 & 12.0      & \xmark & \xmark & \xmark & \xmark & \xmark & \xmark \\ 
Meta R-CNN~\cite{yan2019meta}  & - & 3.5  & - & 2.8  & \xmark & \xmark  & \xmark & \xmark \\ 
MTFA~\cite{ganea2021incremental}  & 2.5 & 6.6  & 2.7 & 6.6  & \xmark  & \xmark & \xmark & \xmark \\ 
iMTFA~\cite{ganea2021incremental}  & 3.3 & 6.2  & 2.8 & 5.2 & \xmark   & \xmark & \xmark & \xmark \\ 
Mask-DeFRCN~\cite{qiao2021defrcn}  & 7.5 & 15.4  & 6.7 & 12.7 & \xmark   & \xmark & \xmark & \xmark \\ 
DCFS~\cite{gao2022decoupling}  & 8.1 & 16.4  & 7.2 & \textbf{13.5} & \xmark & \xmark  & \xmark & \xmark \\ 
Meta-DETR~\cite{zhang2022meta}   & - &  15.3 & - & 8.1 & \xmark & \xmark  & \xmark & \xmark \\ 
DCFS-pose~\cite{gao2022decoupling}  &11.0 &17.1  &\xmark & \xmark  &9.9 &21.1 &\xmark & \xmark  \\
CFOCNet~\cite{yang2021class}  & \xmark & \xmark & \xmark & \xmark     & \xmark & \xmark & 4.43$^\dagger$  & 2.26$^\dagger$ \\ 
MPSR~\cite{wu2020multi}  & 5.1 & 8.7    & \xmark & \xmark & \xmark & \xmark & 1.42$^\dagger$ & 1.40$^\dagger$ \\  
FSDetView~\cite{xiao2020few}  &  4.5 & 10.7    & \xmark & \xmark & \xmark & \xmark & 1.42$^\dagger$ & 1.42$^\dagger$ \\ 
Mask-RCNN-ft~\cite{he2017mask}  & 2.4 & 6.9 & 2.0 & 5.5 & 2.3 & 6.7 & {1.48}$^\dagger$ & 1.45$^\dagger$ \\
\midrule
\rowcolor{ourscolor}
UniFS (Ours) & \textbf{12.7} & \textbf{18.2} & \textbf{8.6} & 11.5 & \textbf{12.2} & \textbf{22.1}  & \textbf{1.38}$^\ddagger$ & \textbf{1.32}$^\ddagger$  \\
\bottomrule
\end{tabular}
}
\label{tab:coco_result}
\end{table*}

\subsection{Experimental Analysis}

\subsubsection{Experiments on PASCAL-5$^i$ benchmark.}
PASCAL-5$^i$ dataset~\cite{everingham2010pascal} is a well-known benchmark dataset for evaluating few-shot object detection. The dataset contains 20 categories, which are randomly divided into three different splits (Novel Set \#1, \#2, \#3), each consisting of 15 base classes and 5 novel classes. 
Following previous works~\cite{qiao2021defrcn,gao2022decoupling}, training is conducted on the PASCAL VOC 2007+2012 train/val sets, and evaluation is conducted on the PASCAL VOC 2007 test set. 
The Average Precision results for novel classes ($AP_{50}$) in K-shot settings are reported.
Table~\ref{tab:voc_results} summarises our results and compares them with the current state-of-the-art few-shot object detectors on PASCAL VOC. We observe that the proposed method outperforms state-of-the-art specialist models.

\begin{table*}[tb] 
\centering
\caption{\textbf{Few-shot object detection results on PASCAL-5$^i$.} We evaluate model performance ($AP_{50}$) on three different splits. The results are averaged over 10 seeds.
}
\setlength{\tabcolsep}{1.2mm}
\scalebox{0.75}{
    \begin{tabular}{l|c|cccccca}
    \toprule
        Method & Shot & YOLO-ft \cite{kang2019few} & Meta R-CNN \cite{yan2019meta} & TFA  \cite{wang2020frustratingly} & TIP \cite{li2021transformation} & DeFRCN  \cite{qiao2021defrcn} & DCFS \cite{gao2022decoupling} & UniFS \\ \hline
        \multirow{2}{*}{Novel Set \#1} & K=1 & 6.6 & 19.9 & 39.8 & 27.7 & 46.2 & 46.2 & \textbf{47.2} \\ 
         & K=5 & 24.8 & 45.7 & 55.7 & 50.2 & 62.4 & 62.9 & \textbf{63.6} \\ 
        \midrule
        \multirow{2}{*}{Novel Set \#2}  & K=1 & 12.5 & 10.4 & 23.5 & 22.7 & 32.6 & 32.6 & \textbf{33.4} \\ 
         & K=5 & 16.1 & 34.8 & 35.1 & 40.9 & 48.3 & 47.9 & \textbf{49.3} \\ 
        \midrule
        \multirow{2}{*}{Novel Set \#3}  & K=1 & 13 & 14.3 & 30.8 & 21.7 & 39.8 & 40.3 & \textbf{41.2} \\ 
         & K=5 & 32.2 & 41.2 & 49.5 & 44.5 & 56.1 & 56.9 & \textbf{58.5} \\ \hline
    \end{tabular}}
\label{tab:voc_results}
\end{table*}

\subsubsection{Effect of multi-task learning.}
In Table~\ref{tab:multitask}, we observed that co-learning with multiple tasks leads to improved overall performance. This enhancement can be attributed to the inter-task synergy that arises from jointly training different instance perception tasks.

\begin{table*}[htb]
\centering
\begin{minipage}[t]{0.47\textwidth}
\centering
\caption{\textbf{Effect of multi-task learning on the COCO-UniFS \texttt{val} set.}
}
\small
\scalebox{0.63}{
\setlength{\tabcolsep}{6pt}
\begin{tabular}{l c c c c c c}
\toprule
\multirow{2}{*}{Model} & \multicolumn{2}{c}{Det. AP $\uparrow$} &  \multicolumn{2}{c}{Seg. AP $\uparrow$} & \multicolumn{2}{c}{Kpt. AP $\uparrow$}  \\
\cmidrule(r){2-3}
\cmidrule(r){4-5}
\cmidrule(r){6-7}
& K=1 & K=5  & K=1 & K=5  & K=1 & K=5    \\
\toprule
\\
Det.    &12.2 &17.9  &  \xmark & \xmark   & \xmark & \xmark   \\
\\
Det.+Seg. &12.6 &17.6 &  8.5 & 11.2   & \xmark & \xmark  \\
\\
Det.+Seg.+Kpt.   & \textbf{12.7} & \textbf{18.2}  & \textbf{8.6} & \textbf{11.5} & \textbf{12.2} & \textbf{22.1}  \\ 
\bottomrule
\end{tabular}
}
\label{tab:multitask}
\end{minipage}
\hspace{1.mm}
\begin{minipage}[t]{0.47\textwidth}
\centering
\caption{\textbf{Effect of Structure-Aware Point Learning (SAPL).}}
\centering
\small
\scalebox{0.63}{
\setlength{\tabcolsep}{4pt}
\begin{tabular}{l c c c c c c}
\toprule
\multirow{2}{*}{Model} & \multicolumn{2}{c}{Det. AP $\uparrow$} &  \multicolumn{2}{c}{Seg. AP $\uparrow$} & \multicolumn{2}{c}{Kpt. AP $\uparrow$}  \\
\cmidrule(r){2-3}
\cmidrule(r){4-5}
\cmidrule(r){6-7}
& K=1 & K=5  & K=1 & K=5  & K=1 & K=5    \\
\toprule
L2 only  &10.9 &16.1  &  6.1 & 7.7 & 9.1 & 19.5   \\
L1 only  &10.6 &16.2  &  7.2 & 8.7 & 12.0 & 21.0   \\ 
L1+1-hop SAPL   &12.6 &17.9  &  8.4 & 11.3   & 12.3 & 21.8  \\
L1+2-hop SAPL  & \textbf{12.7} &\textbf{18.2} & \textbf{8.6} & \textbf{11.5} & 12.2 & \textbf{22.1}  \\
L1+3-hop SAPL   &12.6 &17.7  &  8.2 & 11.0   & \textbf{12.4} & 21.6 \\
L1+4-hop SAPL   &12.7 &17.8  &  8.4 & 11.3   & 12.2 & 20.9  \\

\bottomrule
\end{tabular}
}
\label{tab:sapl}
\end{minipage}

\end{table*}

\subsubsection{Effect of Structure-Aware Point Learning (SAPL).}
In Table~\ref{tab:sapl}, we analyze the effect of SAPL loss, and compare with using L1/L2 loss only. Interestingly, L1 loss is superior to L2 loss on segmentation and keypoint, and achieves similar performance on detection. We also find that adopting SAPL could bring increased performance with accelerated learning convergence by considering the structural information. For example, the segmentation AP increases from 7.2 AP to 8.4 AP. As 1-hop SAPL is easily affected by point noises, incorporating 2-hop SAPL is also beneficial, further increasing the segmentation AP to 8.6 AP. However, 3-hop and 4-hop structural cues are too smooth to capture detailed shape information. Therefore, we choose to use $N=2$ in our implementation.

\begin{figure}[t]
\centering
    \includegraphics[width=0.98\textwidth]{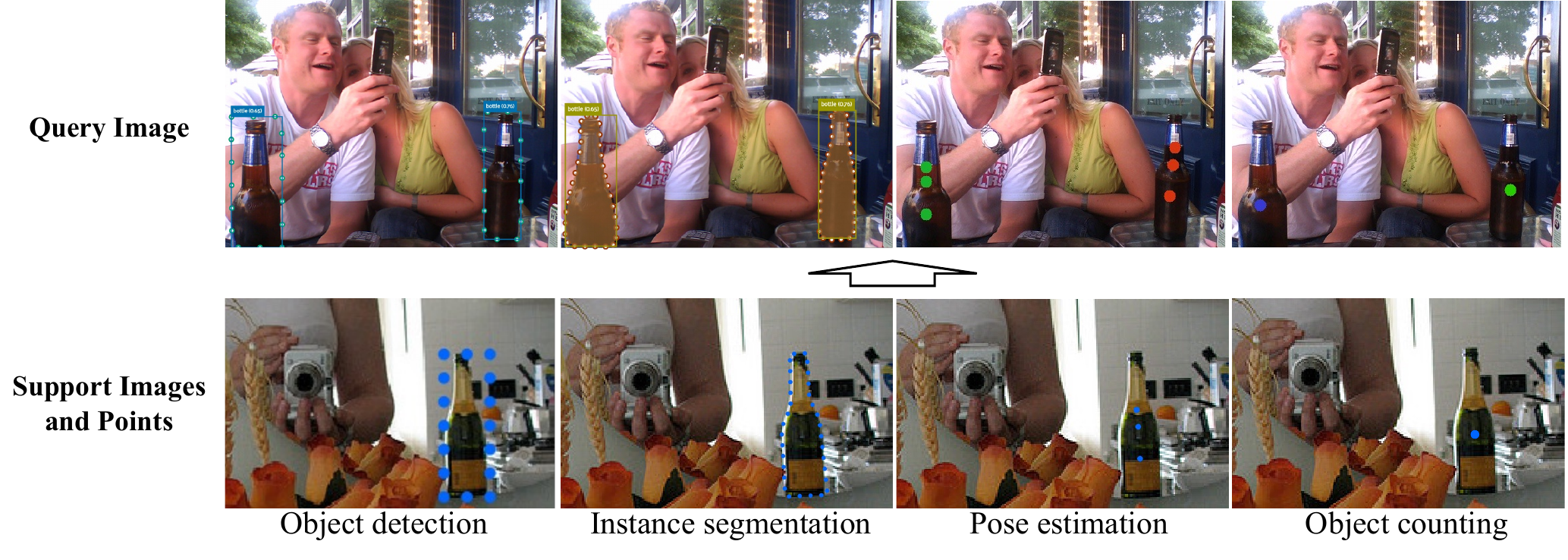}
\caption{Visualization of tasks in a 5-shot scenario with support images of new categories and point annotations, leading to corresponding point outputs on the query set.}
\label{fig:visualization}
\end{figure}

\subsubsection{Qualitative Analysis.}
In Fig.~\ref{fig:visualization}, we qualitatively evaluate the generalization ability of UniFS to novel categories. Our model can follow instructions to perform different tasks with different point annotations as the input. 

\section{Conclusion}
Few-shot learning has been extensively studied to mimic the ability of humans to generalize to new tasks with only a few demonstrations. However, few-shot learning of multi-task instance perception is under explored.
This paper introduces UniFS, a universal few-shot instance perception model that handles a diverse set of instance perception tasks in a unified point representation learning paradigm. 
To facilitate model training and evaluation, we also extend existing datasets and construct the COCO-UniFS benchmark for multi-task few-shot instance perception. 
It is worth noting that our proposed generalist model is achieved with a fully shared model architecture and minimum task-specific designs, which is widely considered to be a unique feature for general artificial intelligence.
We hope for broad applicability and further research on more unified vision models.

\textbf{Limitations and future work.} 
Since point-based method may present challenges for segmentation due to sampling errors, there is room for improvement through further optimization. 
This work primarily focuses on point representation for task unification, and the image-level classification aspect of few-shot learner is not specifically improved.
While our work focuses on 2D tasks, we remain optimistic about its potential for tasks such as 3D or sequential input. Future research endeavors should focus on exploring more comprehensive multi-task perception in few-shot scenarios.

\noindent\textbf{Acknowledgement.}
This paper is partially supported by the National Key R\&D Program of China No.2022ZD0161000 and the General Research Fund of Hong Kong No.17200622 and 17209324.

%
%
\bibliographystyle{splncs04}
\bibliography{egbib}

\clearpage
\appendix

\setcounter{table}{0}
\renewcommand{\thetable}{A\arabic{table}}
\setcounter{figure}{0}
\renewcommand{\thefigure}{A\arabic{figure}}

\setcounter{table}{0}
\renewcommand{\thetable}{S\arabic{table}}
\setcounter{figure}{0}
\renewcommand{\thefigure}{S\arabic{figure}}
\setcounter{section}{0}
\renewcommand{\thesection}{S\arabic{section}}

\section{Effect of Point Number} 
We evaluate the performance of UniFS with different numbers of points to represent the instance perception tasks, as depicted in Table~\ref{tab:effect-point-num}. Initially, we explore the point numbers as 4, 8, and 16 for object detection. It exhibites superior results when applied 16 points, showcasing notably enhanced performance in 5-shot scenario and comparable performance in 1-shot scenario. Therefore, we adopt 16 points for the task of object detection. Subsequently, we conduct experiments on instance segmentation with point numbers as 16, 32, and 64. It demonstrates performance improvement with increased point number on the task of instance segmentation, while introduces a detrimental effect on detection performance. To achieve a trade-off between detection and segmentation, we select to employ 32 points for instance segmentation with decent perforamce on both tasks.

\begin{table*}[tb]
\caption{\textbf{Effect of point number on the COCO-UniFS \texttt{val} set.} The row with blue color indicates our choice for UniFS.
}
\centering
\small
\scalebox{0.98}{
\setlength{\tabcolsep}{10pt}
\begin{tabular}{c c | c c c c}
\toprule
\multirow{2}{*}{Det. Point} &\multirow{2}{*}{Seg. Point} & \multicolumn{2}{c}{Det. AP $\uparrow$} &  \multicolumn{2}{c}{Seg. AP $\uparrow$} \\
\cmidrule(r){3-4}
\cmidrule(r){5-6}
& & K=1 & K=5  & K=1 & K=5  \\
\toprule
4 &-    &{12.4} &17.3  & \xmark   & \xmark  \\
8 &-    &12.2  &16.7 &\xmark   & \xmark\\
16 &-    &12.2 & 17.9  &\xmark   & \xmark\\
\hline
16 &16    & 12.9 &16.5 &  8.6 & 10.7  \\
\rowcolor{ourscolor}
16 &32    &12.6 &17.6 &  8.5 & 11.2  \\
16 &64    &12.0 &17.0 &  9.0 & 11.4  \\
\bottomrule
\end{tabular}
}
\label{tab:effect-point-num}
\end{table*}

\section{Effect of Transformer Blocks} 
We also analyze the influence of the number of transformer blocks applied for the point decoder in UniFS. As shown in Table~\ref{tab:effect-decoder}, we compare the performance of all the tasks with different block numbers. When the number is set to 2, the best results are achieved across most of the tasks. Considering computational cost and model performance, we opt for 2 blocks for point decoder.

\begin{table*}[tb]
\caption{\textbf{Effect of transformer block number on the COCO-UniFS \texttt{val} set.} The row with blue color indicates our choice for UniFS.
}
\centering
\small
\scalebox{0.95}{
\setlength{\tabcolsep}{10pt}
\begin{tabular}{c | c c c c c c c c}
\toprule
\multirow{2}{*}{Number} & \multicolumn{2}{c}{Det. AP $\uparrow$} &  \multicolumn{2}{c}{Seg. AP $\uparrow$} & \multicolumn{2}{c}{Kpt. AP $\uparrow$}  & \multicolumn{2}{c}{Cnt. MSE $\downarrow$} \\
\cmidrule(r){2-3}
\cmidrule(r){4-5}
\cmidrule(r){6-7}
\cmidrule(r){8-9}
& K=1 & K=5  & K=1 & K=5  & K=1 & K=5  & K=1 & K=5   \\
\toprule
1    &12.1 &17.4  &8.0   & 11.0   & 9.6 & 20.6   &1.38 &1.31\\
\rowcolor{ourscolor}
2   & 12.7 & 18.2 & 8.6 & 11.5 & 12.2 & 22.1  & 1.38 & {1.32}  \\
4   &11.9 &17.2  &  8.0 & 11.0 & 10.9 & 20.7 &1.39 & 1.29\\
6   &11.8 &16.9  &  8.2 & 10.9 & 12.6 & 21.6 &1.38 &1.31 \\

\bottomrule
\end{tabular}
}
\label{tab:effect-decoder}
\end{table*}

\begin{figure}[t]
\centering
    \includegraphics[width=0.9\textwidth]{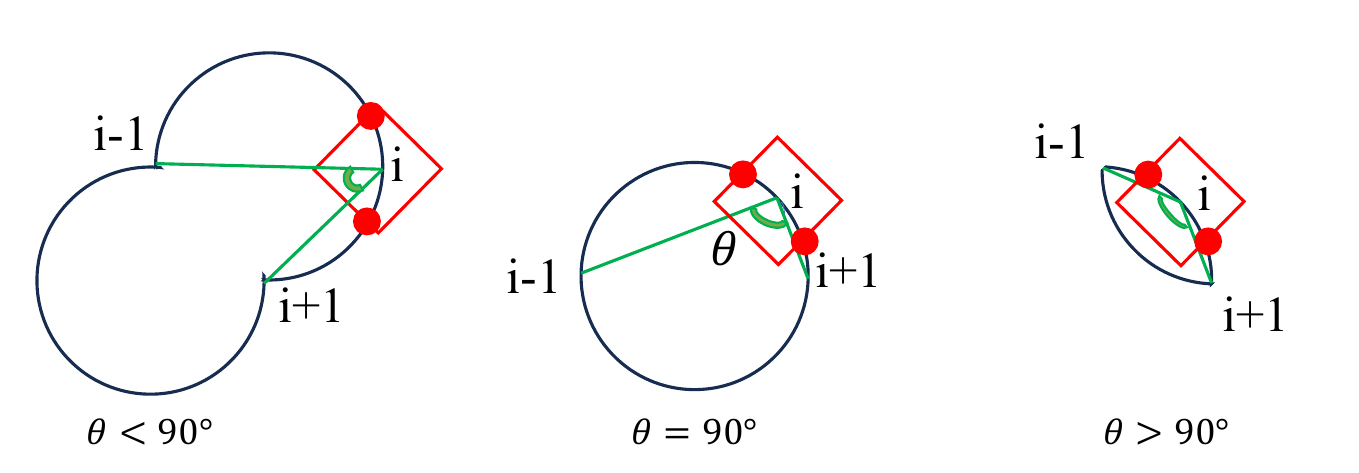}
\caption{Analysis on 1-hop SAPL. For 1-hop SAPL, the trajectory of point P, where the angle between a moving point P and two fixed points is a fixed value $\theta$, is a closed curve composed of two symmetrical arcs: the spindle shape ($\theta<90^\circ$), a circle ($\theta=90^\circ$), and the lens shape ($\theta>90^\circ$). }
\label{fig:analysis-sapl}
\end{figure}

\begin{figure}[t]
\centering
    \includegraphics[width=\textwidth]{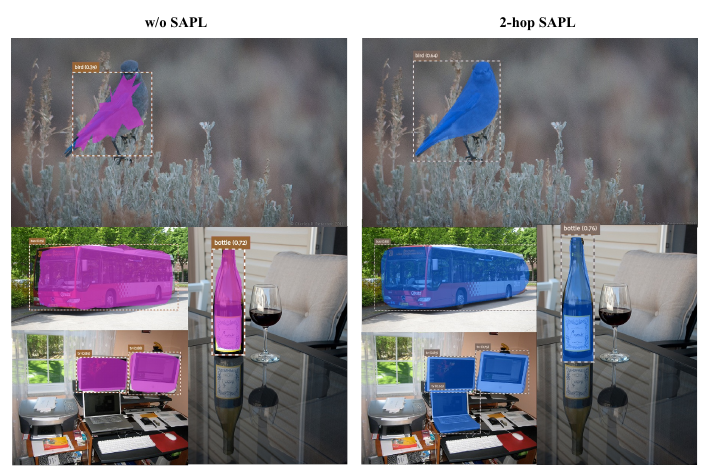}
\caption{\textbf{Qualitative effect of SAPL on the tasks of object detection and instance segmentation.} Our proposed UniFS (``2-hop SAPL'') significantly outperforms the model only using L1 loss (``w/o SAPL'').}
\label{fig:visualization-sapl}
\end{figure}

\section{Analysis on Structure-Aware Point Learning}

In Fig. 3 of the main text , we show that only using L1/L2 loss will introduce ambiguity, as different model predictions have the same loss values (points on the circle for L2, and the diamond for L1). 
As shown in Fig~\ref{fig:analysis-sapl}, for 1-hop SAPL, the trajectory of point P, where the angle between a moving point P and two fixed points is a fixed value $\theta$, is a closed curve composed of two symmetrical arcs: the spindle shape ($\theta<90^\circ$), a circle ($\theta=90^\circ$), and the lens shape ($\theta>90^\circ$). 
Considering both L1 and SAPL, the same loss value occurs at the intersection of their trajectories (two red points as shown below), significantly mitigating the ambiguity.

Table 5 in the main text demonstrates a significant improvement on the tasks of object detection and instance segmentation by introducing Structure-Aware Point Learning (SAPL). We further provide some visualization results to visually illustrate the effectiveness of our proposed SAPL.
As shown in Fig~\ref{fig:visualization-sapl}, it is obvious that UniFS with SAPL effectively learns the object stuctures and produces reliable contours, while the predictions of model trained with L1 loss only (``w/o SAPL'') are inaccurate. For example, in the topmost image, the model without SAPL cannot capture the relationship between points when supervising each point individually, leading to unsmooth boundary and incorrect bounding box.

\end{document}